\documentclass[10pt,twocolumn,letterpaper]{article}

\usepackage{cvpr}
\usepackage{times}
\usepackage{epsfig}
\usepackage{graphicx}
\usepackage{amsmath}
\usepackage{amssymb}

\usepackage{booktabs}
\usepackage{multirow}
\usepackage{algorithm}
\usepackage{algorithmic}
\usepackage{mathrsfs} 
\usepackage{units}
\usepackage{xfrac}

\cvprfinalcopy 

\def\cvprPaperID{3127} 

\ifcvprfinal\fi
\begin{document}

\title{Network Adjustment: Channel Search Guided by FLOPs Utilization Ratio}

\author{Zhengsu Chen\textsuperscript{1},\quad Jianwei Niu\textsuperscript{1},\quad Lingxi Xie\textsuperscript{2},\quad Xuefeng Liu\textsuperscript{1},\quad Longhui Wei\textsuperscript{3},\quad Qi Tian\textsuperscript{4}\\
\textsuperscript{1}Beihang University,\quad\textsuperscript{2}Johns Hopkins University,\quad\textsuperscript{3}Peking University,\quad\textsuperscript{4}Xidian University\\
{\tt\small danczs@buaa.edu.cn},\quad{\tt\small niujianwei@buaa.edu.cn},\quad{\tt\small 198808xc@gmail.com},\\
{\tt\small liu\_xuefeng@buaa.edu.cn},\quad{\tt\small longhuiwei@pku.edu.cn},\quad{\tt\small wywqtian@gmail.com}
}

\maketitle
\thispagestyle{empty}

\begin{abstract}
Automatic designing computationally efficient neural networks has received much attention in recent years. Existing approaches either utilize network pruning or leverage the network architecture search methods. This paper presents a new framework named \textbf{network adjustment}, which considers network accuracy as a function of FLOPs, so that under each network configuration, one can estimate the \textbf{FLOPs utilization ratio} (FUR) for each layer and use it to determine whether to increase or decrease the number of channels on the layer. Note that FUR, like the gradient of a non-linear function, is accurate only in a small neighborhood of the current network. Hence, we design an iterative mechanism so that the initial network undergoes a number of steps, each of which has a small `adjusting rate' to control the changes to the network. The computational overhead of the entire search process is reasonable, i.e., comparable to that of re-training the final model from scratch. Experiments on standard image classification datasets and a wide range of base networks demonstrate the effectiveness of our approach, which consistently outperforms the pruning counterpart. The code is available at \url{https://github.com/danczs/NetworkAdjustment}.

\end{abstract}

\section{Introduction}

In the deep learning era~\cite{lecun2015deep}, as researchers designed more and more complicated neural networks~\cite{krizhevsky2012imagenet,simonyan2014very,he2016deep,huang2017densely,zoph2017learning} for various computer vision problems~\cite{girshick2014rich,long2015fully}, it becomes increasingly important to find efficient architectures that are friendly to real-world applications. The computational overhead of a deep neural network is closely related to its width, \textit{i.e.}, the number of channels on each layer of it. Beyond manually designed networks which have fixed configurations of width~\cite{he2016deep,simonyan2014very,szegedy2015going}, researchers have also explored the direction of automatically determining the numbers of channels. Representative methods include network pruning~\cite{li2016pruning,liu2017learning,huang2018data} and channel search~\cite{tan2019mnasnet, cai2018proxylessnas}, both of which obtained success in improving the computational efficiency (accuracy with respect to overhead) of the network. 
However, both of these two approaches have some limitations. Network pruning methods usually start with a wide network and try to reduce the channel number of each layer. These methods are mostly not sensitive to specific efficiency criteria (\textit{e.g.}, FLOPs) so that the computational resource is not always assigned to the places that need it most. On the other hand, the main drawbacks of the network search techniques are the computational complexity and search strategy which limit the search space and increase the search cost.



In this paper, we propose a novel pipeline named \textbf{network adjustment} which starts with a pre-trained network, and simultaneously adds channels to some layers and subtracts channels from some others, with the criterion of the utilization ratio of resources at each layer.
Network adjustment stands out from existing approaches in two aspects. \textbf{First}, we can estimate the state of each individual convolutional layer and optimize the channel numbers simultaneously. Therefore, our method is very efficient for channel search. \textbf{Second}, we search the channel configuration with an efficiency-aware strategy, by which we can find computationally efficient networks.
The top-level design of our approach is an iterative process, in which each iteration involves training the given network, computing the utilization ratio on each layer, and adjusting the channel numbers accordingly. The computational overhead of the entire process is often comparable to that of training the initial network from scratch, yet we can observe gradual improvements of computational efficiency throughout iterations.

As a particular example, we aim to optimize the FLOPs (floating point operations) of deep networks, and calculate the \textbf{FLOPs utilization ratio} (FUR) for this purpose. FUR is defined on \textbf{each layer} of a network, measuring the accuracy gain that can be obtained by increasing unit FLOPs on that layer. To compute it, we assume that network accuracy is a function of FLOPs and that FUR, like the gradient of a non-linear function, can be estimated with an acceptable error in a small neighborhood of each state. To mimic a small amount of FLOPs change, we propose a randomized approach which discards channels with a small drop probability (also named as SpatialDropout~\cite{tompson2015efficient}) in each layer individually, and approximate FUR to be the ratio of accuracy drop over FLOPs drop, assuming that SpatialDropout removes the equivalent number (may be a floating point number) of channels. Note that FUR computation only involves network inference and thus executes efficiently.

We evaluate network adjustment on the standard benchmark of designing efficient networks. On both the CIFAR-100 and ImageNet datasets, our approach consistently improves the accuracy of a few popular network backbones, with the FLOPs barely changed. Compared with state-of-the-art structured pruning or channel search algorithms, network adjustment also enjoys advantages in terms of recognition accuracy under the same FLOPs.

The remainder of this paper is organized as follows. Section~\ref{related_work} briefly reviews the previous literature, and Section~\ref{approach} elaborates the network adjustment approach. Experiments are shown in Section~\ref{experiments} and conclusions are drawn in Section~\ref{conclusions}.

\section{Related Work}
\label{related_work}

The number of channels in each layer is an important factor in designing deep neural networks. Researchers started with empirical methods. A popular example is VGGNet~\cite{simonyan2014very} which increases the channel number each time when the spatial resolution is reduced. This strategy was followed by a lot of work with either manual~\cite{xie2016aggregated,zoph2017learning, szegedy2015going,szegedy2016inception} or automatic architecture designs~\cite{zoph2017learning,real2019regularized, liu2018darts, pham2018efficient}. There also exist methods that increased the channel number gradually with the layer index, such as PyramidNet~\cite{han2017deep} and DenseNet~\cite{huang2017densely}. To reduce the computational burden, researchers proposed a `bottleneck' module~\cite{he2016deep} which reduces the number of channels before a costly computation and recovers it thereafter. Nevertheless, such work requires considerable manual efforts and the designed architectures were believed imperfect in many aspects.

To explore a larger number of possibilities, the idea of neural architecture search (NAS) was introduced~\cite{zoph2017learning,wu2019fbnet}, so that the number of channels at each layer is determined with an automatic algorithm guided by general heuristics, \textit{e.g.}, the computational efficiency of the network. A typical example is to optimize a RNN controller to control the strategy of sampling architectures from the large search space~\cite{tan2019mnasnet, howard2019searching}. However, the computational burden largely limits the search space of such methods. For acceleration, researchers had to reduce the number of candidates to be used in each layer~\cite{cai2018proxylessnas}, or trained a wide network beforehand and sampled channels from each layer to mimic the behavior of networks with different widths~\cite{guo2019single,chen2019detnas,stamoulis2019single}. Both strategies introduced inaccuracy to the network, which leads to sub-optimal search results. The differentiable methods~\cite{liu2018darts, chen2019progressive, xu2019pc} have not been applied to channel search yet.

Another line of research lies in channel pruning, which aims to speed up the network by removing a part of channels which contribute little to recognition~\cite{li2016pruning,liu2017learning,huang2018data}. The typical pipeline of pruning starts with a well-trained network, on which various criteria are used to measure the importance of a neuron~\cite{han2015learning,han2015deep}, a channel~\cite{li2016pruning,liu2017learning}, or a layer~\cite{wen2016learning,wang2018skipnet}. After that, less important units are removed and the network gets fine-tuned~\cite{li2016pruning}. Recently, researchers started rethinking the value or working mechanism of network pruning~\cite{liu2018rethinking,frankle2018lottery}, which led to novel designs of pruning approaches~\cite{yu2019network,lym2019prunetrain}.

\section{Our Approach}
\label{approach}

\begin{figure*}[!t]
\centering
\includegraphics[width=2.0\columnwidth,trim=0 100 0 120,clip]{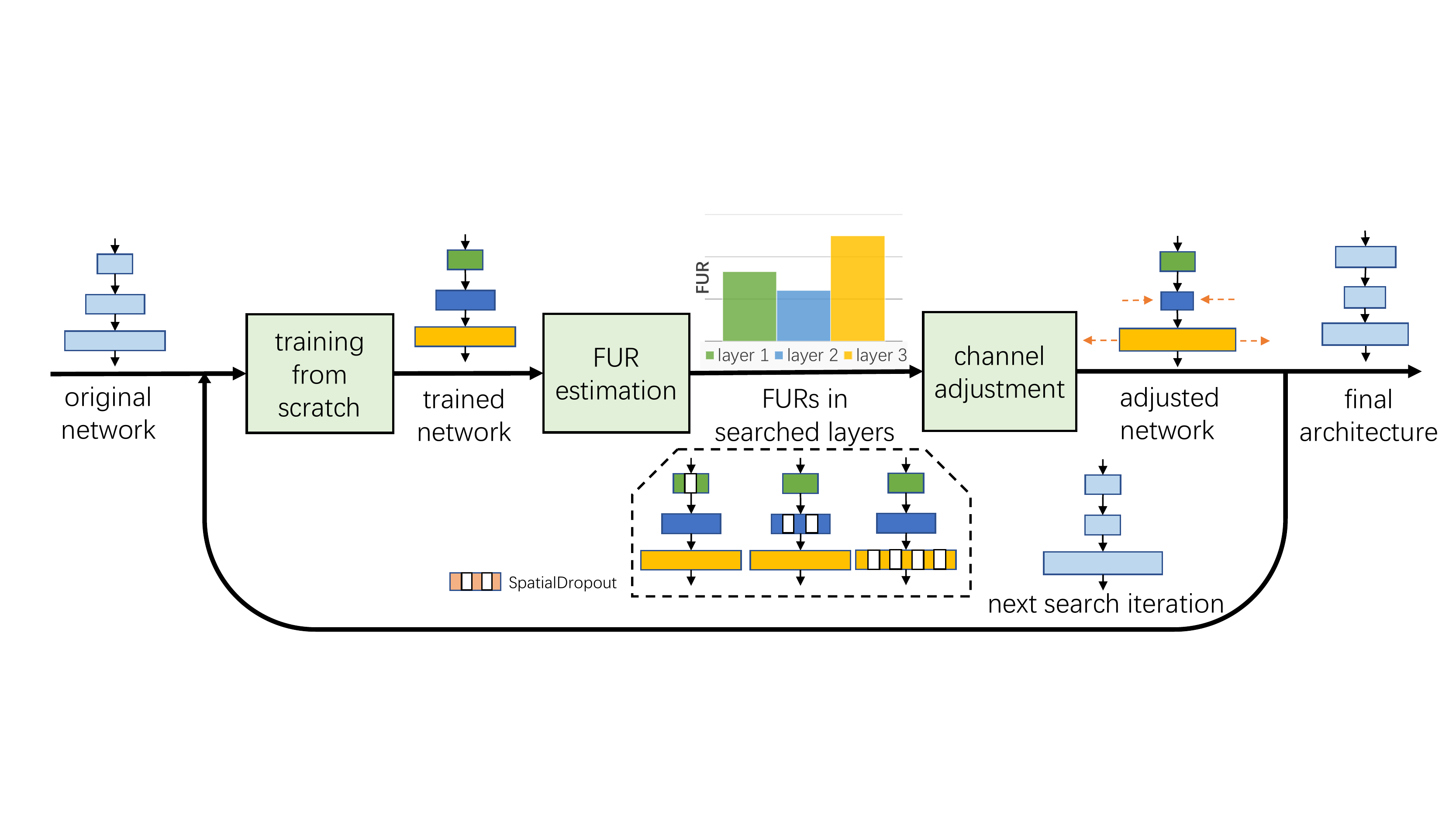} 
\caption{
The framework of our channel number search approach (best viewed in color). The leftmost shows the original network, which undergoes an iterative process. In each round, the current network is trained from scratch, after which we estimate the FUR of each layer (marked in different colors) by individually performing SpatialDropout~\cite{tompson2015efficient} on each layer. A taller bar indicates a higher FUR. After that, we choose the layers (only 1 in this example) with the highest FUR and increase its width, and do the opposite thing for the layers with the lowest FUR. This iteration continues till convergence, or a pre-defined number of iterations has been arrived.}
\label{framework}
\end{figure*}

\subsection{Problem}

%
%
%

Given a network $\mathbb{N}$ with $L$ convolutional layers, a channel configuration of $\mathbf{c}$ =  ($c_{1}$, $c_{2}$, ..., $c _{L}$), and a FLOPs function, $\mathrm{FLOPs}(\mathbf{c})$, the goal is to improve the network performance by adjusting the channel configuration with pre-defined FLOPs, $F_0$:
\begin{align}
 \label{defination}
&\mathbf{c}^\star = \arg\mathop{\max}\limits_{\mathbf{c}}\mathrm{Acc}_{\mathrm{val}}(\mathbf{c},\mathbf{W}^\star) \\
&\mathrm{s.t.}\quad\mathrm{FLOPs}(\mathbf{c}) = F_0,\quad\mathbf{W}^\star = \arg\mathop{\max}_{\mathbf{W}}\mathrm{Acc}_\mathrm{train}(\mathbf{c},\mathbf{W}) \nonumber
\end{align}
where $\mathbf{W}$ is the parameters of the network and $F_0$ is the FLOPs of the initial network. $\mathrm{Acc}_\mathrm{train}$ and $\mathrm{Acc}_\mathrm{val}$ denote the training and validation accuracy, respectively.

The problem is the channel numbers $\mathbf{c}$ are discrete integers for a network and cannot be solved by optimization algorithms directly. To solve $\mathbf{c}$ like gradient based optimization algorithms, we, therefore, propose a method to estimate the gradients of them, which will be detailed in Section~\ref{sec_fur}. The estimated gradients are also named as FLOPs utilization ratios or FLOPs utilization efficiency in this paper. The gradients of $\mathbf{c}$ are positive in general because most networks will benefit from increasing channels. To ensure $\mathrm{FLOPs}(\mathbf{c}) = F_0$, we increase the channels in the layers with higher FURs and reduce the same amount of FLOPs from the layers with lower FURs. The channel adjustment will be executed iteratively like gradient based optimization algorithms until the FURs are similar in every layer.

\subsection{Pipeline: Network Adjustment}

Our method aims to search for more efficient channel configuration by considering the efficiency more accurately. During search, the channel numbers are adjusted iteratively. There are three steps within each iteration: network training, FUR estimation and channel adjustment as shown in Figure~\ref{framework}. 
\textbf{First}, we optimize the current network on the training set. This process starts from scratch, but the number of epochs can be significantly smaller than a regular, complete training process because we only need to determine the FLOPs utilization efficiency of each layer, which does not necessarily mean that the network has arrived at a high accuracy. \textbf{Second}, we compute the FLOPs utilization ratio on each layer with the method described in the next part. This process only involves sampling-based inference on the validation set, and is thus relatively fast. \textbf{Third} and last, we use a pre-defined adjusting rate (similar to learning rate) to increase the numbers of channels of a few layers with top-ranked FURs, and decrease the numbers of a few layers with bottom-ranked FURs. The FLOPs of the entire network must remain close to the original amount after each iteration, if not, we use a scale function to slightly increase or decrease the channel numbers of all layers~\cite{gordon2018morphnet}. A typical search process involves a few (\textit{e.g.}, 10) iterations, or it can be terminated if the accuracy is saturated on the validation set.

The pseudo code is summarized in Algorithm~\ref{alg:pipeline}. The validation set is usually split from the original training set. The function `$\mathrm{Scale}$' scales the channel numbers to keep the FLOPs unchanged after adjustment. After search, the network with configuration $\mathbf{c}^\star$ will be trained on the full training set and evaluated on the testing set.

\vspace{0.2cm}
\noindent$\bullet$\quad\textbf{Difference from Prior Work}

Before continuing with the technical part, we elaborate the difference between our approach and two families of prior work. The difference between our method and \textbf{network pruning} mainly lies in two aspects. First, pruning methods mostly focus on finding the less important channels and evaluate it with absolute performance drop~\cite{yu2019network} or reconstruction loss~\cite{huang2018data}. In contrast, our method considers the channels in a layer as a whole and evaluates the layer with FUR. Second, our method adjusts the channel configuration finely and iteratively while most pruning methods obtain the pruned model from a pre-trained wide model directly. 

Our method is also different from \textbf{channel search} methods. Although many NAS systems update the channel numbers iteratively~\cite{zoph2017learning,tan2019mnasnet}, their methods usually need hundreds of GPU days to converge. That is because their search systems can only update channel numbers according to the network performance. In our method, we evaluate the FUR in each layer to guide the search process and the search system can converge after a few iterations (\textit{e.g.}, 10). Additionally, limited by memory and computational resource, most NAS systems only search configuration from a few channel number candidates. Our method, however, can learn the channel configuration with an increased degree of freedom.

\begin{algorithm}[t!]
\caption{Network Adjustment} 
\textbf{Input:} training set $\mathcal{D}_\mathrm{train}$, validation set $\mathcal{D}_\mathrm{val}$, initial channel configuration $\mathbf{c}^{\left(1\right)}$, adjusted layers $K$, adjusting rate $r_\mathrm{A}$, max iterations $T$;\\
\textbf{Output:} adjusted channel configuration $\mathbf{c}^\star$;

\begin{algorithmic}[1]
\FOR{$t=1,2,\ldots,T$}
\STATE Build a network $\mathbb{N}$ with  $\mathbf{c}^{\left(t\right)}$;
\STATE Train the network $\mathbb{N}$ on the training set $\mathcal{D}_\mathrm{train}$;
\STATE Test the network $\mathbb{N}$ and obtain the accuracy $\mathrm{Acc}_\mathrm{val}^{\left(t\right)}$ on the validation set $\mathcal{D}_\mathrm{val}$;
\STATE Compute the FUR for each layer of $\mathbb{N}$ on the validation set: $\mathbf{F}^{\left(t\right)} = (\mathrm{FUR}_{1}^{\left(t\right)},\mathrm{FUR}_{2}^{\left(t\right)},...,\mathrm{FUR}_{L}^{\left(t\right)})$;
\STATE Sort the elements in $\mathbf{F}^{\left(t\right)}$ and obtain a top index set, $\mathcal{I}_\mathrm{top}^{\left(t\right)}$, with $K$ largest elements in $\mathbf{F}^{\left(t\right)}$, and a bottom set $\mathcal{I}_\mathrm{bot}^{\left(t\right)}$, with $K$ smallest elements in $\mathbf{F}^{\left(t\right)}$;
\STATE  $\mathbf{c}^{t+1} = \mathbf{c}^{t}$;
\FOR{$i$ in $\mathcal{I}_\mathrm{top}^{\left(t\right)}$}
\STATE  $c_{i}^{\left(t+1\right)} = \mathrm{Round}(c_{i}^{\left(t\right)} + r_\mathrm{A} \cdot c_{i}^{\left(1\right)})$;
\ENDFOR
\FOR{$i$ in $\mathcal{I}_\mathrm{bot}^{\left(t\right)}$}
\STATE  $c_{i}^{\left(t+1\right)} = \mathrm{Round}(c_{i}^{\left(t\right)} - r_\mathrm{A} \cdot c_{i}^{\left(1\right)})$;
\ENDFOR
\STATE  $\mathbf{c}^{\left(t+1\right)} = \mathrm{Scale}(\mathbf{c}^{\left(t+1\right)})$;
\ENDFOR
\STATE  $t^\star = \arg\mathop{\max}\limits_{t}(\mathbf{Acc}_\mathrm{val}^{\left(t\right)})$;
\STATE  $\mathbf{c}^\star = \mathbf{c}^{\left(t^\star\right)}$;
\end{algorithmic}
\textbf{Return:} $\mathbf{c}^\star$.
\label{alg:pipeline}
\end{algorithm}

\subsection{FLOPs Utilization Ratio}
\label{sec_fur}

Following the above, the \textbf{FLOPs utilization ratio} (FUR) of a layer is defined to be the contribution that a unit amount of FLOPs makes to the network accuracy, \textit{i.e.}, ${\partial\mathrm{Acc}_\mathrm{val}(\mathbf{c},\mathbf{W}^\star)}/{\partial\mathrm{FLOPs}(\mathbf{c})}$. To compute this quantity, we make use of the channel numbers, $\mathbf{c}$, as an intermediate variable, and compute two quantities. The \textbf{first} quantity is the gradient of $\mathrm{Acc}_\mathrm{val}(\mathbf{c},\mathbf{W}^\star)$ with respect to $\mathbf{c}$, denoted as ${\partial\mathrm{Acc}_\mathrm{val}(\mathbf{c},\mathbf{W}^\star)}/{\partial\mathbf{c}}$. In particular, an element in this vector, ${\partial\mathrm{Acc}_\mathrm{val}(\mathbf{c},\mathbf{W}^\star)}/{\partial c_l}$, indicates the contribution that a unit channel makes to the final accuracy. The \textbf{second} quantity involves $\mathrm{FLOPs}(\mathbf{c})$, and we use another vector, $\mathbf{f}$, which shares the same dimensionality with $\mathbf{c}$ and each element, $f_l$, denotes the FLOPs in the $l$-th layer. We have  $\mathrm{FLOPs}(\mathbf{c})=\sum_{l=1}^Lf_l+\mathrm{const}$, where the constant is often determined by the architecture topology. For a convolutional layer, $f_l$ is proportional to the spatial resolution, input and output channel numbers, and the convolutional kernel size of the $l$-th layer. Similarly, computing ${\partial\mathrm{FLOPs}(\mathbf{c})}/{\partial c_l}$ obtains the contribution that a unit channel makes to the resource usage. With these quantities, the FUR of the $l$-th layer is defined as:
\begin{equation}
\mathrm{FUR}_l = \frac{\partial\mathrm{Acc}_\mathrm{val}(\mathbf{c},\mathbf{W}^\star)}{\partial\mathrm{FLOPs}(\mathbf{c})}=\frac{{\partial\mathrm{Acc}_\mathrm{val}(\mathbf{c},\mathbf{W}^\star)}/{\partial c_l}}{{\partial\mathrm{FLOPs}(\mathbf{c})}/{\partial c_l}}
\end{equation}
Here we have omitted the notation of the network topology, $\mathbb{N}$, in all terms, for simplicity.

In writing ${\partial\mathrm{Acc}_\mathrm{val}(\mathbf{c},\mathbf{W}^\star)}/{\partial c_l}$ and ${\partial\mathrm{FLOPs}(\mathbf{c})}/{\partial c_l}$, we assume that $c_l$, the number of channels in each layer, can take continuous (non-integer) values. To mimic this behavior, we introduce a probabilistic mechanism which involves adding SpatialDropout~\cite{tompson2015efficient} to each layer individually, and observing how the network accuracy is affected by this operation. SpatialDropout is proposed as a regularization method and randomly discards channels with a certain probability. To our best knowledge, it is the first time that SpatialDropout is used for channel evaluation. 

To reduce the error in computing FUR, we discard the same FLOPs in every layer. In other words, the change of FLOPs is the same for all layers, by which we can rank the FURs with the accuracy drop directly. Since the data drop probability is continuous, we can simulate dropping fractional channels to guarantee that the discarded FLOPs in all layers are the same.

\subsection{Towards a Slightly Different Network}
\label{sec_adj}

FUR is like the gradient of network performance with respect to FLOPs and is very likely to be accurate only for the current network. Therefore, like the gradient based optimization method, we only slightly adjust the channel numbers in each iteration, so that FURs are still accurate enough. 

Specifically, we increase the channels of the top-$k$ layers ranked by FURs and reduce the channels in the bottom-$k$ layers. The hyperparameter $k$ will be deceased during search. The `adjusting rate' is like the learning rate in gradient based optimization algorithms and is used to adjust the channels in an iteration. It should not be too big (We use 0.1 for most of our experiments). Big adjusting rates may make the search system unstable.
 
When the channel configuration is updated, the FUR in each layer will change. To obtain new FUR of the new network, we will retrain the new networks from scratch. 
Since evaluating a trained network on the validation set is relatively fast, the most time-consuming step in our framework is training a network. In every searching iteration, a network should be trained. For tasks like training big models on ImageNet, this strategy becomes impractical. There are two methods that can tackle this problem:
dataset sampling~\cite{wu2019fbnet} and reducing the training epochs~\cite{tan2019mnasnet}. In our experiment, it is observed that reducing the training epochs has less effect on FUR estimation than dataset sampling. When training for much fewer epochs, the FUR ranking can still be revealed, even though the network performance is relatively low. Hence, we train our model on ImageNet for about ten percent of the full epochs. In this way, the time for search channel numbers on ImageNet is comparable to training a model on ImageNet.

\subsection{Implementation Details}

\textbf{SpatialDropout and network training.}
As mentioned above, we evaluate the FUR by randomly discarding some channels (SpatialDropout). To make the evaluation reasonable, we should reduce unexpected changes (\eg the change of data distribution) caused by discarding channels as much as possible, so that the test accuracy can reveal the real effect caused by channel change. Generally speaking, discarding some channels will not sharply degrade the network performance. The network accuracy usually decreases smoothly as the drop probability increases. In fact, even discarding whole layers in residual blocks will not disable a residual network~\cite{huang2016deep, veit2016residual}. 

Moreover, we use some tricks to make the neural layers accustomed to data drop. Firstly, Spatial Dropout is used during network training. Since the network is trained with SpatialDropout, it will not be sensitive to SpatialDropout during test. 
Secondly, the data distribution is balanced after SpatialDropout like standard Dropout. For example, if $10\%$ of the channels are discarded, the remaining channels will be scaled by $10/9$. Finally, we reduce the disharmony between BN and data drop. There exists disharmony between BN and dropout~\cite{li2019understanding}. It is demonstrated that Gaussian dropout and using dropout after BN can solve this problem~\cite{li2019understanding}.

\begin{figure}[!t]
\centering
\includegraphics[width=0.9\columnwidth,trim=240 60 190 72,clip]{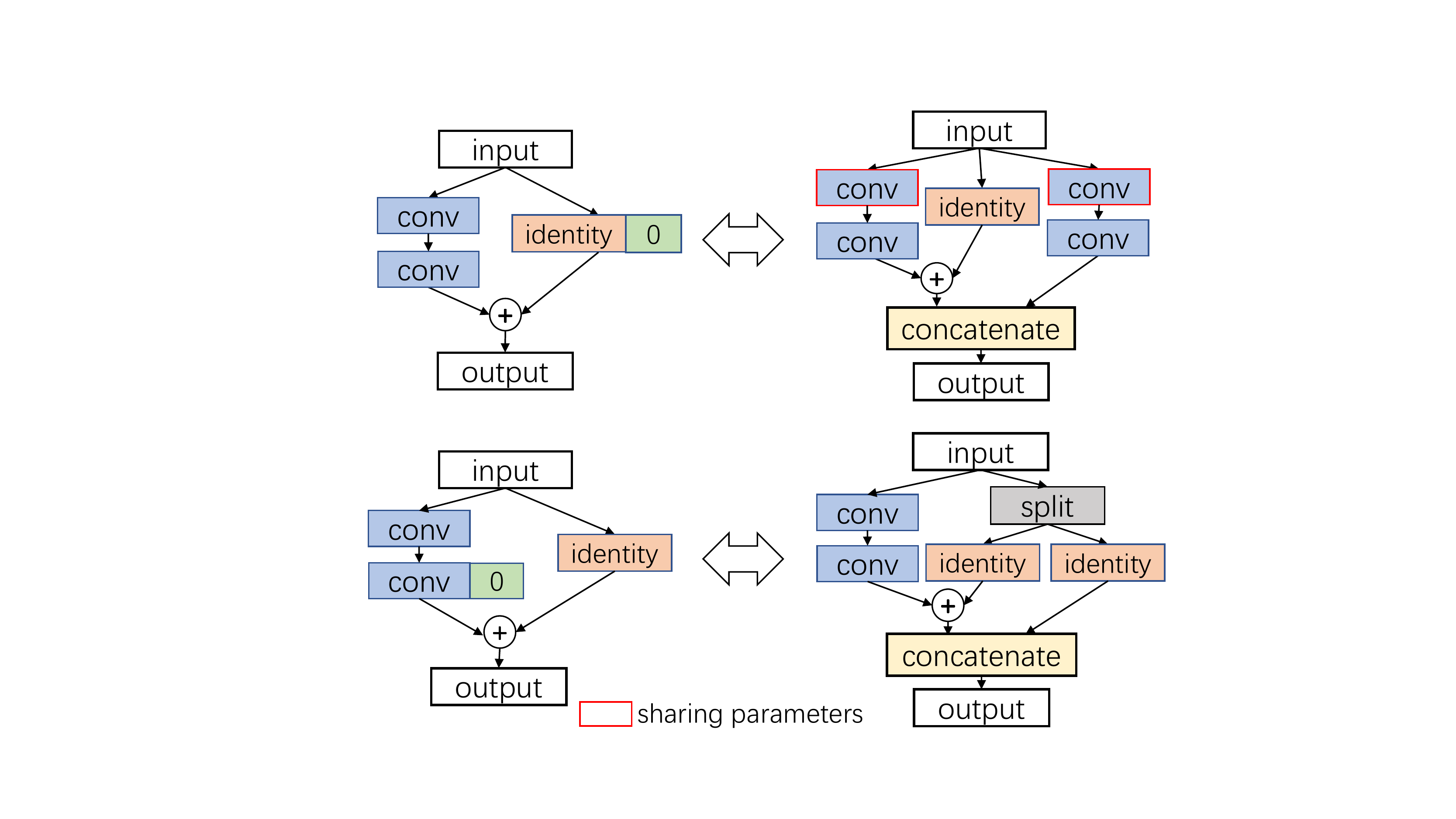} 
\caption{
The equivalent structure when utilizing zero padding. Padding zeros to the path with fewer channels is equivalent to introducing a new path to the residual block.}
\label{zeropad}
\end{figure}

\textbf{Zero padding for neural layers.}
For single-path networks like VGGNets, the channel numbers in every layer can be set freely. For multi-path networks like ResNets, there are some constraints on setting channel numbers. Taking residual blocks as an example, since the output of convolutions should be added to the residual shortcut, the channel numbers of these two paths should be the same. To solve this problem, we simply pad zeros to the path with fewer channels. Zero padding not only enables us to set the channel numbers freely in multi-path networks but also introduces some network structure implicitly. As displayed in Figure~\ref{zeropad}, in residual blocks, padding zeros to shortcut is equivalent to add a convolution path between input and output~\cite{han2017deep}. Padding zeros to convolution output is equivalent to concatenate an extra shortcut to the output. Therefore, searching channel numbers for multi-path networks with zero padding is also searching the network structure implicitly.

\section{Experiments}
\label{experiments}
\subsection{Results on CIFAR-100}
We test our method on CIFAR-100, since the results on CIFAR-100 are more stable than those on CIFAR-10. CIFAR-100 is a publicly available dataset proposed by ~\cite{krizhevsky2009learning}. The dataset consists of 32$\times$32 colored natural images that fall in 100 classes. There are 50,000 images in the training set and 10,000 images in the testing set. We randomly sample 5,000 images from the training set as the validation set to search the channel numbers. Afterwards, we retrain the searched networks on the whole training set and release the results on the testing set. Data augmentation is conducted during training following the common practice~\cite{he2016deep, zagoruyko2016wide}. During search and test, the networks are trained for 200 epochs and SpatialDropout is used for convolutional layers. The learning rate is dropped from 0.15 to 1e-3 with cosine annealing.

\begin{table}[!t]
\newcommand{\tabincell}[2]{\begin{tabular}{@{}#1@{}}#2\end{tabular}}

  \caption{Accuracy (\%) comparison on CIFAR-100 between our approach and other competitors. Here, `D' indicates the network depth, `FT' indicates that the searched network is only fine-tuned, not trained from scratch, and `KD' is for knowledge distillation. The mark `imp' indicates our own implementation. Legend: LCCL~\cite{dong2017more}, SFP~\cite{he2018soft}, FPGM~\cite{he2019filter}, TAS~\cite{dong2019network}.}
  \setlength{\tabcolsep}{0.16cm}
  \label{tab_pruning_nets}
  \centering
  \small
  \begin{tabular}{p{0.35cm}p{2.2cm}p{0.3cm}p{0.3cm}lc}
    \toprule
     
\tabincell{c}{D} & Method& FT & \tabincell{c}{KD}& \tabincell{c}{~~Acc (\%)}&	\tabincell{c}{FLOPs\\ (M)}\\

\midrule
\multirow{8}{*}{20}
&LCCL	&  &&64.66	(-2.87)&27.3\\
&SFP &&&	64.37 (-3.25)&	24.3\\
&FPGM &&&	66.86 (-0.76)&	24.3\\
&TAS &&$\checkmark$&68.90 (+0.21)& 22.4\\
\cline{2-6}
&original (imp)	&&&69.52	&40.1\\
&original 0.75$\times$	&&&65.90	&22.7\\
&searched 0.75$\times$ 	&&&69.04 (-0.48)	&22.4\\
&searched 0.75$\times$ 	&&$\checkmark$&70.03 (+0.51)	&22.4\\

     \midrule
     
     \multirow{10}{*}{32}
&LCCL	&&&67.39	(-2.69)&43.2\\
&SFP &&&	68.37 (-1.40)&40.3\\
&FPGM &&&	68.52 (-1.25)&	40.3\\
\cline{2-6}
&TAS &&&	68.95 (-1.66)&	42.5\\
&TAS &$\checkmark$&&	69.70 (-0.91)&	42.5\\
&TAS &&$\checkmark$&	72.41 (+1.80)&	42.5\\
\cline{2-6}
&original (imp)	&&&72.57	&68.4\\
&original 0.75$\times$ 	&&&69.68	&38.5\\
&searched 0.75$\times$ 	&&&71.93 (-0.64)	&42.2\\
&searched 0.75$\times$ 	&&$\checkmark$&73.82 (+1.25)	&42.2\\

\midrule
\multirow{7}{*}{110}
&LCCL	&  &&70.78	(-2.01)&173\\
&SFP &$\checkmark$&&	71.28 (-2.86) &	121\\
&FPGM &$\checkmark$&&72.55 (-1.59)&	121\\
&TAS &&$\checkmark$&73.16 (-1.90)& 120\\
\cline{2-6}
&original (imp)	&&&75.68	&252\\
&original 0.7$\times$	&&&73.46	&119\\
&searched 0.7$\times$ 	&&&74.84 (-0.84)	&120\\
&searched 0.7$\times$ 	&&$\checkmark$&75.96 (+0.28)	&120\\

\bottomrule
  \end{tabular}
\end{table}

\begin{table*}[!t]
  \caption{The accuracy (on CIFAR-100) and FLOPs of ResNet-20, throughout a complete iterative process of network adjustment.}
  \label{tab_cifar100_iter}
  \setlength{\tabcolsep}{0.16cm}
  \centering
  \small
  \begin{tabular}{lccccccccccc}
    \toprule
       iteration & 0 (original network)  & 1& 2 & 3 & 4  & 5& 6 & 7 & 8  & 9 &10\\
    \midrule
        
  accuracy (\%) & 69.52	&70.12 	&70.24	&70.34 &70.48	&70.94	&71.09&	71.55&	71.47&	71.36 & 71.57\\
  \midrule
FLOPs (M) &40.1	&40.5	&39.8	&40.2	&40.1	&40.3	&40.4	&40.3	&40.3	&40.4 & 40.2\\
    \bottomrule
  \end{tabular}
\end{table*}

\begin{figure*}[!t]
\centering
\includegraphics[width=1.8\columnwidth,trim=20 160 20 130,clip]{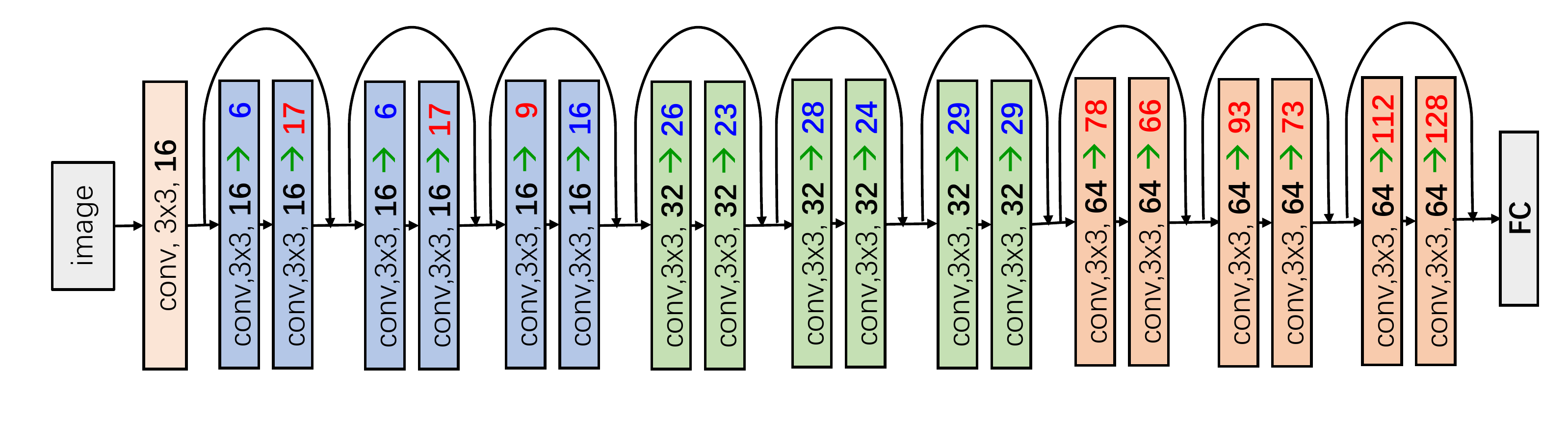} 
\caption{
The configuration of ResNet-20 before and after the entire network adjustment process (10 rounds) on CIFAR-100. Red and blue indicate increased and decreased channel numbers, respectively.}
\label{resnet20}
\end{figure*}

\vspace{0.2cm}
\noindent$\bullet$\quad\textbf{Comparison to Network Pruning}

The results of our method and some pruning methods are shown in Table~\ref{tab_pruning_nets}. To compare our method to the pruning methods, we first narrow down the original networks to a comparable FLOPs level and then search for channel configuration based on the thin networks. 

For network testing, the pruning models are usually initialized with the weights of the original model (fine-tuning), which is usually better than training from scratch. Some methods transfer the knowledge in original models by knowledge distillation~\cite{hinton2015distilling}, which usually can significantly improve the pruned networks. Since we aim to search for better channel configuration, our models are mainly evaluated by training from scratch. We also report the knowledge distillation results for a fair comparison. The methods in Table~\ref{tab_pruning_nets} have different baselines. Thus the main criterion is the accuracy drops of the pruning networks compared to their original networks. As can be seen, our searched models outperform most of the pruning methods on accuracy drop. For ResNet-32, although the TAS model with knowledge distillation performs better than our model, our search network outperforms the TAS model when trained from scratch, which demonstrates that our channel configuration is better. For ResNet-110, our searched model that is trained from scratch even performs better than the TAS model with knowledge distillation.

\vspace{0.2cm}
\noindent$\bullet$\quad\textbf{Diagnostic Studies}
%

\textbf{Different iterations on CIFAR-100}. The channel number search results of ResNet-20 on CIFAR-100 is shown in Table~\ref{tab_cifar100_iter}. As can be seen, the performance of the network is consistently improved in the first seven iterations. Afterwards, the performance gain becomes saturated. Since we have balanced the increase and decrease of FLOPs in different layers, the FLOPs of the network are almost unchanged. 
Note that the parameters of the networks are increased. It reveals that the network is exploring the unconstrained resources to improve its performance. A similar phenomenon is observed in~\cite{chu2019fairnas}. We think the exchange of parameters and performance can be a new direction of designing networks. Because for most current devices, FLOPs rather than parameters are the bottleneck.

\begin{table}[!t]
\newcommand{\tabincell}[2]{\begin{tabular}{@{}#1@{}}#2\end{tabular}}
  \caption{The searched results for different networks.}
  \label{tab_cifar100_nets}
  \centering
  \small
  \begin{tabular}{lllc}
    \toprule
      
  Networks & method &	\tabincell{c}{Accuray\\ (\%)}&\tabincell{c}{FLOPs \\(M)}\\
     \midrule
     \multirow{2}{*}{ResNet-20}&original	&69.52	&40.1\\ 
&searched	&71.57 (+2.05)	&40.2\\
  \midrule
  
     \multirow{2}{*}{ResNet-32}&original	&72.57	&68.4\\
&searched	&74.14 (+1.57)	&68.2\\
   \midrule
   
     \multirow{2}{*}{ResNet-56}&original	&74.90	&125\\
&searched	&76.15 (+1.25)	&126\\
%
%
   \midrule
  \multirow{2}{*}{ResNet-110}&original	&75.68&	252 \\
&searched&76.98 (+1.30)	&254	\\
   \midrule
     \multirow{2}{*}{ResNet-20 2.0$\times$}&original	&75.60	&160\\
&searched	&76.87 (+1.27)	&161\\
   \midrule      
   
\multirow{2}{*}{ResNet-20 4.0$\times$}&original	&78.84	&642\\ 
&searched&79.51 (+0.67)&	641	\\
  \midrule
  \multirow{2}{*}{Plain-20}&original&65.62	&40.1\\
&searched	&68.89 (+3.27)	&40.1\\
  \midrule
\multirow{2}{*}{DenseNet-40}&original&76.32	&282\\
&searched&76.95 (+0.63)&282\\
\midrule
\multirow{2}{*}{PyramidNet-20}&original&66.58	&40.4\\
&searched &71.53 (+4.95)	&40.1\\

    \bottomrule
  \end{tabular}
\end{table}

\textbf{The searched architecture}. The searched channel configuration in ResNet-20 is different from the original network both at network-level and substructure-level. The structure of the original ResNet-20 and searched ResNet-20 are shown in Figure~\ref{resnet20}. At network-level, more channels are assigned to the deep layers. It demonstrates that these layers are more efficient in FLOPs utilization. At substructure-level, as for the layers that closed to the input, a more efficient residual block structure is found. In the first residual block, the output channel number of the first layer decreases from 16 to 6. Afterwards, the channel number is increased to 17 by the second convolutional layer. Although the output channel numbers of the block are slightly increased compared to the original block, the FLOPs of the residual block are sharply reduced. In this residual block, reducing the output of the first convolution will decrease the FLOPs of the two convolutional layers at the same time. Reducing the output of the second convolution may only decrease the FLOPs of the second layer, because the output will be added to the residual connection and the FLOPs of the following layer will not change if its channels are fewer than the residual connection. Our method is sensitive to FLOPs utilization and, thus, can find this `compression-expansion' structure. The result also indicates that our method can not only reset the channel numbers at network-level but also find more efficient substructure in neural networks. 

\textbf{Without FUR}. We study FUR by searching ResNet-20 without it. It means that when evaluating channels, we randomly drop a fixed ratio of channels in each layer and rank the layers by the accuracy drops instead of FURs. Under this setting, the searched ResNet-20, which achieves 71.06\%, still outperforms the original network (69.52\%), but performs worse than the network searched with FUR (71.57\%). At substructure-level, the fixed-ratio network can not learn the `compression-expansion' structure shown in Figure~\ref{resnet20}, because it is not aware of the difference of FLOPs utilization between the two convolutional layers. 

\textbf{Different initialization of channel numbers}. To test the effect of the initialization channel numbers, we construct a 20 layer PyramidNet~\cite{han2017deep}. The channels of this network are linearly increased and the FLOPs are similar to ResNet-20. With the same FLOPs, ResNet-20 outperforms PyramidNet-20 by 2.94\%. From the perspective of FLOPs, PyramidNet-20 is a bad initialization of channel configuration for a 20 layer residual network. After channel search, the two searched networks achieve similar performance as shown in Table~\ref{tab_cifar100_nets}. It indicates that our method can overcome the bad initialization of channel numbers.


\textbf{Different depths, different widths, different architectures.}
We test our model on different width and depth ResNets. When increasing depth, our method still works well. ResNet-32, 56 and 110 outperform the original networks by over 1\%. For these deep networks, more FLOPs are assigned to the last several layers like ResNet-20. In ResNet-56 and 110, the `compression-expansion' structure is learned not only in the first stage but also in the last network stage. ResNet-20 2.0$\times$ and ResNet-20 4.0$\times$ are networks widened from ResNet-20 by 2.0 and 4.0. These wide networks can also benefit from channel adjustment.

Plain-20 is the non-residual version of ResNet-20. The searched result on Plain-20 outperforms the original network by 3.27\%. The searched channel configuration of DenseNet-40 is different from ResNet-20. More FLOPs are assigned to 28th-36th layers rather than the last several layers. It indicates that the FURs in these layers are higher than the others for the original DenseNet-40.

\textbf{Channel configuration transfer.}
To evaluate the generalization of the searched channel configuration, the searched configuration of ResNet-20 is transferred to ResNet-20 2.0$\times$ by doubling the channels. The transferred model has the same FLOPs with ResNet-20 2.0$\times$ and the accuracy is 76.34\% on CIFAR-100. Although the transferred model outperforms the original model, it performs worse than the directly searched model (76.87\%). It indicates that there are biases when transferring channel configuration among networks with different widths.

\subsection{Results on ImageNet}

\begin{table*}[thp!]
\caption{Top-1 accuracy (\%) comparison on ImageNet between our approach and other competitors. Here, `FT' indicates that the searched network is only fine-tuned, not trained from scratch, and `KD' is for knowledge distillation. The mark `imp' indicates our own implementation. Legend: LCCL~\cite{dong2017more}, SFP~\cite{he2018soft}, FPGM~\cite{he2019filter}, TAS~\cite{dong2019network}, AutoSlim~\cite{yu2019network}.}
\newcommand{\tabincell}[2]{\begin{tabular}{@{}#1@{}}#2\end{tabular}}
\label{tab_imagenet}
\centering
\small
\begin{tabular}{llccllcc}
\hline
Networks & Method& FT & KD & Top-1 (\%)& Top-5 (\%)&	FLOPs (M)&Parameters(M)\\  
\hline
\multirow{8}{*}{ResNet-18}&LCCL	&&&66.33 (-3.65) &86.94 (-2.29)	&1.19E3&-\\
&SFP &&&67.10 (-3.18)& 87.78 (-1.85)&1.06E3&	-\\
&FPGM &$\checkmark$&&68.41 (-1.87)&	88.48 (-1.15)&	1.06E3&	-\\
&TAS &&$\checkmark$&69.15 (-1.5)&	89.19 (-0.68)&	1.21E3&	-\\
\cline{2-8}
&original (imp)	&&&70.31&89.45	&1.81E3&	11.7\\
&searched	&&&71.22 (+0.91)&90.07 (+0.62)	&1.81E3&	18.5\\
&searched 0.8$\times$ 	&&&69.41 (-0.90)&88.71 (-0.74)	&1.17E3&	11.8\\
&searched 0.8$\times$ 	&&$\checkmark$&71.00 (+0.69)&90.05 (+0.60)	&1.17E3&	11.8\\
\hline
\multirow{3}{*}{MnasNet}&original (imp)	&&&74.21&91.83	&312&	3.89\\
\cline{2-8}
&AutoSlim &&&74.60 (+0.60)&-&	315&6.00	\\
\cline{2-8}
&ours	&&&74.88 (+0.67) &92.15 (+0.32)	&312&	4.96\\
\hline
\end{tabular}
\end{table*}

\begin{figure*}[t!]
\centering
\includegraphics[width=1.8\columnwidth,trim=5 170 35 130,clip]{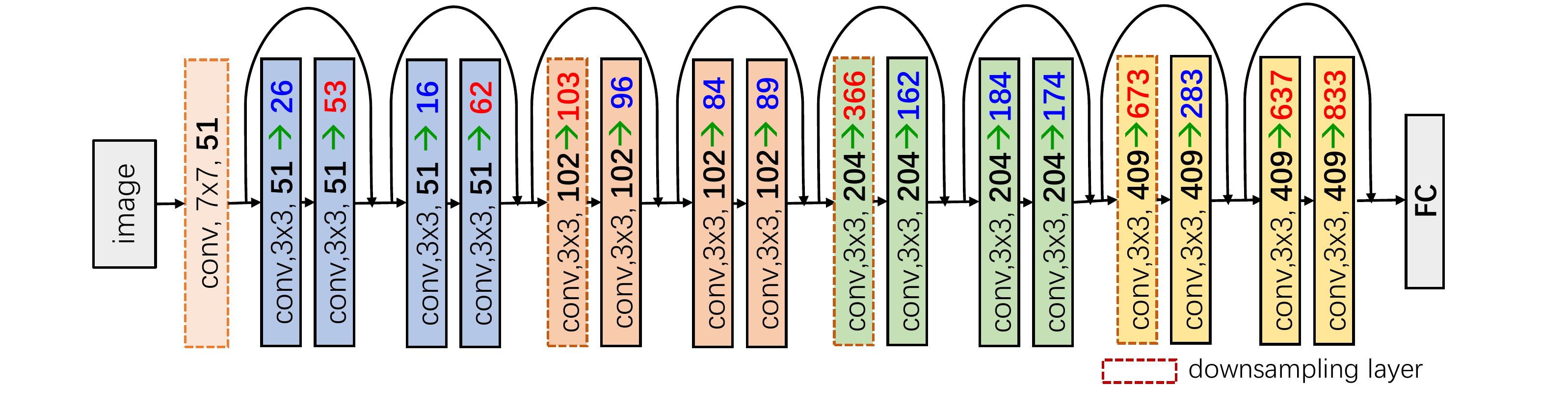} 
\caption{
The configuration of ResNet-18 0.8$\times$ before and after the entire network adjustment process (10 rounds) on ImageNet. Red and blue indicate increased and decreased channel numbers, respectively.}
\label{resnet18}
\end{figure*}

\setcounter{figure}{4}
\begin{figure}[t]
\begin{center}
\includegraphics[width=0.9\linewidth]{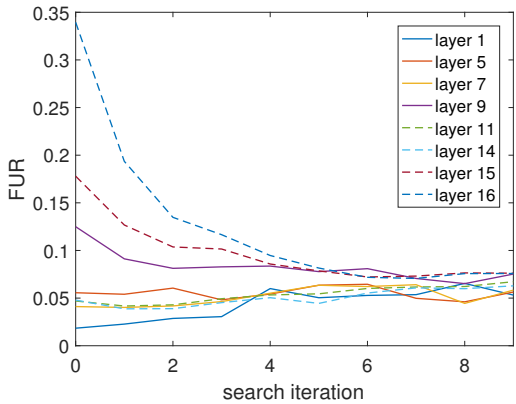}
\end{center}
\vspace{-0.4cm}
\caption{FURs of convolutional layers in ResNet-18 $0.8\times$. We sampled 8 out of 16 layers for better visualization.}
\label{fig:FURplot}
\end{figure}

ILSVRC2012 is a subset of ImageNet database~\cite{russakovsky2015imagenet}. There are 1.3M, 50K and 150K images in training, validation and testing set. The images fall into 1K categories. We randomly sample 50K images from the training set as the searching validation set. Afterwards, we retrain the searched networks on the whole training set and release the results on the original validation set. The images are augmented with the method in~\cite{szegedy2016rethinking}. On ImageNet, networks are trained for 10 epochs in each iteration and the channel number of the first convolutional layer is fixed during search. During the re-training process, ResNet-18 is trained on the full training set for 90 epochs. The learning rate is decayed from 0.1 to 0.001 with cosine annealing. MnasNet is trained for 500 epochs and the learning rate is dropped from 0.075 to 0.0001.

\textbf{First}, we compare our results to those produced by network pruning. For ResNet-18, the searched model outperforms the original model by 0.91\% for top-1 accuracy as shown in Table~\ref{tab_imagenet}. To compare our method to pruning approaches, we scale the channel numbers in ResNet-18 by a factor of 0.8 and search for channel configuration based on it. The top-1 accuracy of ResNet-18 0.8$\times$ outperforms the other pruning approaches on accuracy drop and with knowledge distillation, our model performs much better.

The original and searched ResNet-18 0.8$\times$ are shown in Figure~\ref{resnet18}. As can be seen, the first stage also learned the `compression-expansion' structure and more FLOPs are assigned to the last several layers. In addition, the downsampling layers in the networks are assigned more channels, indicating that the downsampling layers are different from other layers and should be specifically designed. We also show the FURs of some layers during the adjustment process in Figure~\ref{fig:FURplot}, from which one can observe how the FURs differ from each other in the beginning and gradually become similar during the adjustment.

\textbf{Second}, we adjust beyond MnasNet~\cite{tan2019mnasnet}, a searched network architecture based on MobileNetV2. By changing the channel configuration, we improve the network performance with the same FLOPs, as shown in Table~\ref{tab_imagenet}. It indicates that our method can find the more efficient channel configuration compared to a NAS approach in a limited search space. Our method slightly outperforms AutoSlim with fewer parameters. Like ResNet-18 0.8$\times$, the searched MnasNet also assigns more FLOPs to the downsampling layers, which indicates that the current structure of downsampling can be further improved.

\section{Conclusions}
\label{conclusions}

This paper presents a new pipeline named network adjustment for designing efficient network architectures. Our approach is motivated by the idea of measuring how FLOPs are utilized in a pre-trained model, so that it is possible to adjust the number of network channels according to the efficiency of computational resource utilization. For this purpose, we consider network performance as a function of FLOPs and its `gradient', FLOPs utilization ratio (FUR), can be estimated and utilized in a small neighborhood. Integrating all the above yields an iterative pipeline, which outperforms state-of-the-art network pruning methods in terms of recognition accuracy under the same amount of computation. Our research puts forward a new point that network accuracy and speed should be optimized jointly, for which network adjustment provides a preliminary solution. More efforts are required along this path.


\vspace{0.2cm}
\noindent
\textbf{Acknowledgements}\quad
This work was supported by the National Key R\&D Program of China (2017YFB1301100), National Natural Science Foundation of China (61772060, U1536107, 61472024, 61572060, 61976012, 61602024), and the CERNET Innovation Project (NGII20160316). 

{\small
\bibliographystyle{ieee_fullname}
\bibliography{egbib}
}

\end{document}